\documentclass[letterpaper, 10 pt, conference]{ieeeconf}  

\IEEEoverridecommandlockouts                              

\usepackage{amsmath,amssymb,amsfonts}
\usepackage{algorithmic}
\usepackage{graphicx}
\usepackage{textcomp}
\usepackage{xcolor}
\usepackage{hyperref}
\usepackage{svg}
\usepackage{threeparttable} 
\usepackage{booktabs}
\usepackage{array}
\usepackage{longtable}
\usepackage{subcaption}
\usepackage{float}
\usepackage{fancyvrb}
\usepackage{fvextra}
\usepackage{tcolorbox}
\usepackage{lipsum} 
\usepackage{listings}
\usepackage{soul}
\usepackage{fancyhdr}
\usepackage{titlesec}

\newcolumntype{L}{>{\centering\arraybackslash}m{3cm}}
\usepackage[linesnumbered,ruled,vlined]{algorithm2e}

\title{\LARGE \bf
HyperGraph ROS: An Open-Source Robot Operating System for Hybrid Parallel Computing based on Computational HyperGraph
}

\author{Shufang Zhang, Jiazheng Wu, Jiacheng He, Kaiyi Wang, Shan An$^{\ast}$, \textit{Senior Member, IEEE} 
\thanks{This work was supported by the National Key Research and Development Program of China (Grant No. 2023YFC3603601) and Tianjin Research Innovation Project for Postgraduate Students (2022SKYZ367)} 
\thanks{Shufang Zhang, Jiazheng Wu, Jiacheng He, Kaiyi Wang, and Shan An are with the School of Electrical and Information Engineering, Tianjin University, Tianjin 300072, China. {\tt\small\{shufangzhang, wujiazheng, hejiacheng, kaiyiwang, anshan\}@tju.edu.cn}} 
\thanks{$^{\ast}$Corresponding author.}
}

\begin{document}

\fancypagestyle{firstpage}{
    \fancyhf{} 
    \fancyhead[L]{\small This work has been submitted to the IEEE for possible publication. Copyright may be transferred without notice, after which this version may no longer be accessible.}
    \renewcommand{\headrulewidth}{0pt} 
}

\setlength{\headsep}{10pt} 

\maketitle
\thispagestyle{firstpage}

\begin{abstract}
This paper presents HyperGraph ROS, an open-source robot operating system that unifies intra-process, inter-process, and cross-device computation into a computational hypergraph for efficient message passing and parallel execution. In order to optimize communication, HyperGraph ROS dynamically selects the optimal communication mechanism while maintaining a consistent API. For intra-process messages, Intel-TBB Flow Graph is used with C++ pointer passing, which ensures zero memory copying and instant delivery. Meanwhile, inter-process and cross-device communication seamlessly switch to ZeroMQ. When a node receives a message from any source, it is immediately activated and scheduled for parallel execution by Intel-TBB. The computational hypergraph consists of nodes represented by TBB flow graph nodes and edges formed by TBB pointer-based connections for intra-process communication, as well as ZeroMQ links for inter-process and cross-device communication. This structure enables seamless distributed parallelism. Additionally, HyperGraph ROS provides ROS-like utilities such as a parameter server, a coordinate transformation tree, and visualization tools. Evaluation in diverse robotic scenarios demonstrates significantly higher transmission and throughput efficiency compared to ROS 2. Our work is available at \url{https://github.com/wujiazheng2020a/hyper_graph_ros}.
\end{abstract}

\section{Introduction}
Building a complete robotic system requires developers to possess strong programming skills, particularly in parallel computing. However, even with such expertise, the process of developing, debugging, and maintaining a robotic system incurs significant overhead. To address this challenge, ROS (Robot Operating System)~\cite{quigley2009ros,macenski2022robot} encapsulates essential software components required for building robotic systems into libraries. These libraries include cross-process and cross-device communication modules based on the publisher-subscriber mechanism, as well as predefined usage conventions. Additionally, ROS provides various debugging and visualization tools, which significantly reduce the development time and maintenance costs of robotic systems. As a result, ROS (both ROS 1 and ROS 2) has become widely adopted in robotic applications.

\begin{figure}[thpb]
    \centering
    \vspace{0.2cm} 
    \includegraphics[scale=0.115]{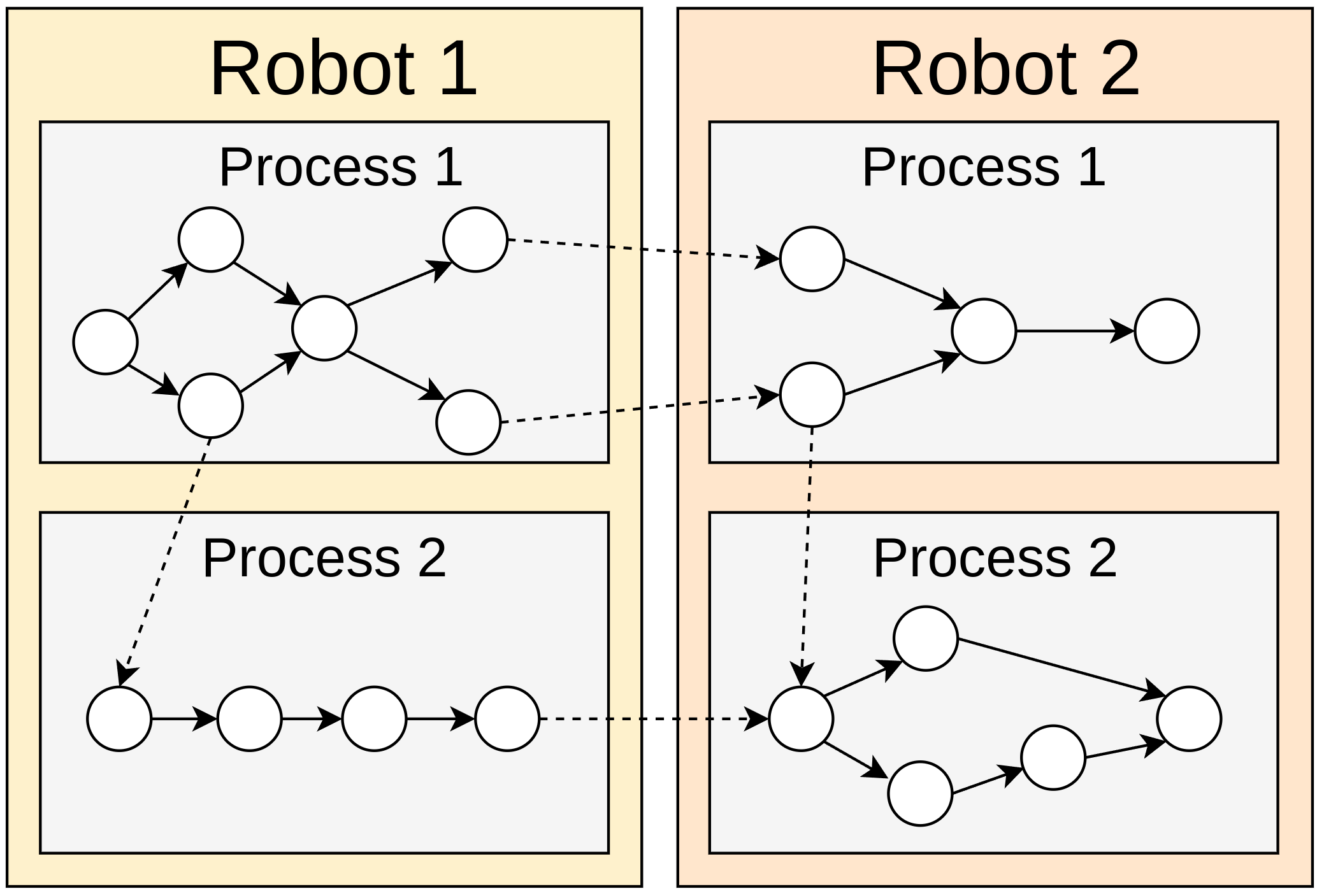}
    \caption{Overview of the proposed HyperGraph. Circles represent Intel-TBB \texttt{function\_node}s, each executing data processing or algorithms. Solid arrows denote edges between \texttt{function\_node}s within a process, forming an internal computational graph. Dashed arrows represent ZeroMQ edges, connecting the internal computational graphs of each process across different robots or processes, thereby constructing an overarching HyperGraph.}
    \label{intro_picture}
\end{figure}
However, practical use of ROS presents several limitations. In intra-process communication, ROS 1 uses \texttt{Nodelet} for message passing, while ROS 2 enhances this with pointer-based communication enabled by the \texttt{use\_intra\_process\_comms} flag. However, this flag affects the functionality of debugging tools and monitoring nodes. Therefore, dynamically adjusting the communication mechanism without modifying the API represents a key improvement. Secondly, after receiving messages, nodes in ROS lack an efficient parallel scheduling mechanism, which could be a potential enhancement for high real-time robotic applications. 
Third, both ROS 1 and ROS 2 rely on their own serialization protocol, necessitating additional memory copies for algorithms based on Protocol Buffers (protobuf)~\cite{protobuf_cite} or custom protocols. This leads to excessive memory and computational overhead, particularly in large-scale rendering and visualization tasks.
Fourth, although ROS 1 and ROS 2 offer an RViz-based visualization debugging interface~\cite{kam2015rviz}, it is often too resource-intensive for real-time robotic debugging. In practice, while RViz is just one of many debugging tools, its memory and computational resource consumption is disproportionately high compared to its actual importance in debugging.

To address these issues, we introduce the following improvements in the proposed operating system:
\begin{itemize}
    \item \textbf{High-Performance Unified API for Intra-Process, Inter-Process, and Cross-Device Communication}: Developers can configure message transmission through a parameter file without modifying the code. If no external communication is specified, messages are passed directly via C++ pointers between intra-process nodes, incurring \textbf{near-zero memory overhead, computational cost, and latency}.
    
    \item \textbf{Automated Scheduling and Dynamic Parallelism}: When a node receives a message—whether from an external process or another robot—it is automatically integrated into its local Intel TBB Flow Graph~\cite{pheatt2008intel}. The work-stealing algorithm in TBB dynamically allocates resources based on workload distribution, optimizing task parallelism and balancing computational efficiency. This approach enhances parallel efficiency, achieves adaptive load balancing, maximizes CPU utilization, and minimizes synchronization overhead.

    \item \textbf{Efficient Serialization and Deserialization for Cross-Process and Cross-Device Communication}: Since many community applications rely on protobuf, this OS adopts protobuf instead of ROS’s custom serialization. It supports multiple languages, including but not limited to C++, Python, C\#, Go, and Java, ensuring broad compatibility while eliminating redundant memory copies. Moreover, it supports custom serialization methods, further enhancing the efficiency and versatility of our operating system.
    
    \item \textbf{High Frame Rate and Low Memory Overhead Visualization Tool}: The visualization system is built using native OpenGL~\cite{shreiner2009opengl} with GLFW~\cite{glfw_cite} and IMGUI~\cite{dear_imgui_cite}. Compared to the use of Qt~\cite{QtFramework} and OGRE3~\cite{ogre3d} in ROS RViz, this approach achieves a lower memory footprint and higher rendering efficiency. Although it may compromise graphical aesthetics, it significantly improves performance, which is preferable for debugging tools.
    
\end{itemize}

In summary, this OS integrates intra-process, inter-process, and cross-device communication into a unified API, forming a computational hypergraph, as illustrated in Fig. \ref{intro_picture}. It employs distinct mechanisms for intra-process and inter-process messaging while allowing external transmission configuration via parameter files. Additionally, it provides a more efficient visualization tool and auxiliary components, including a coordinate transformation tree, parameter server, time synchronization system, hypergraph configuration tools, etc. Designed for resource-constrained applications such as autonomous vacuum cleaners, patrol robots, drones, etc. The HyperGraph ROS is lightweight and highly efficient. Compared to ROS, it reduces hardware requirements while enhancing computational and communication efficiency. This makes it an ideal choice for embedded and cost-sensitive robotic systems. Finally, it is fully open-source and licensed under the MIT license.

\section{Related Work}
\label{related_work_sec}
Early frameworks such as CARMEN~\cite{montemerlo2003perspectives}, Player~\cite{gerkey2001most}, and RTM~\cite{ando2005rt} introduced modular architectures and message-passing mechanisms that enabled distributed robotic applications. Subsequent middleware systems further refined these concepts: OROCOS~\cite{bruyninckx2003real} focused on real-time control; LCM~\cite{huang2010lcm} and YARP~\cite{metta2006yarp} provided high-performance messaging; and RT-Middleware for VxWorks~\cite{ikezoe2010openrt} targeted embedded environments.

ROS 1~\cite{quigley2009ros} relies on a centralized master node, which means that the entire system loses communication if the master node fails. To address these limitations, ROS 2~\cite{maruyama2016exploring} adopted a decentralized architecture based on the Data Distribution Service (DDS)~\cite{pardo2003omg}. This change introduced multi-threaded executors and quality-of-service controls. However, DDS-based communication still incurs latency and resource overhead in high-frequency, time-sensitive applications~\cite{teper2022end, puck2021performance}. Additionally, its multi-threaded executor may not always guarantee deterministic execution in complex multi-robot scenarios~\cite{dust2023experimental, jo2022smartmbot}.

HyperGraph ROS builds on these insights by employing a hybrid communication model. Specifically, it uses direct pointer passing for intra-process messaging and ZeroMQ~\cite{hintjens2013zeromq, kang2020evaluating} for inter-process and cross-device communication. Additionally, HyperGraph ROS leverages Intel Threading Building Blocks (TBB) for automated scheduling and dynamic parallelism~\cite{kim2010multicore, marowka2012tbbench, newburn2011intel, karcher2014autotuning, saule2012early}. This integration achieves improved load balancing, reduced latency, and enhanced computational efficiency. These design choices directly address the limitations of earlier systems, making HyperGraph ROS well suited for resource-constrained and real-time robotic applications.

\section{System Design}
\subsection{Design Goal}
The proposed system aims to address the limitations of ROS and provide a more efficient and user-friendly robotic operating system. The primary design goals are as follows:
\begin{figure*}[thpb]
    \centering
    \vspace{0.1cm} 
    \includegraphics[scale=0.148]{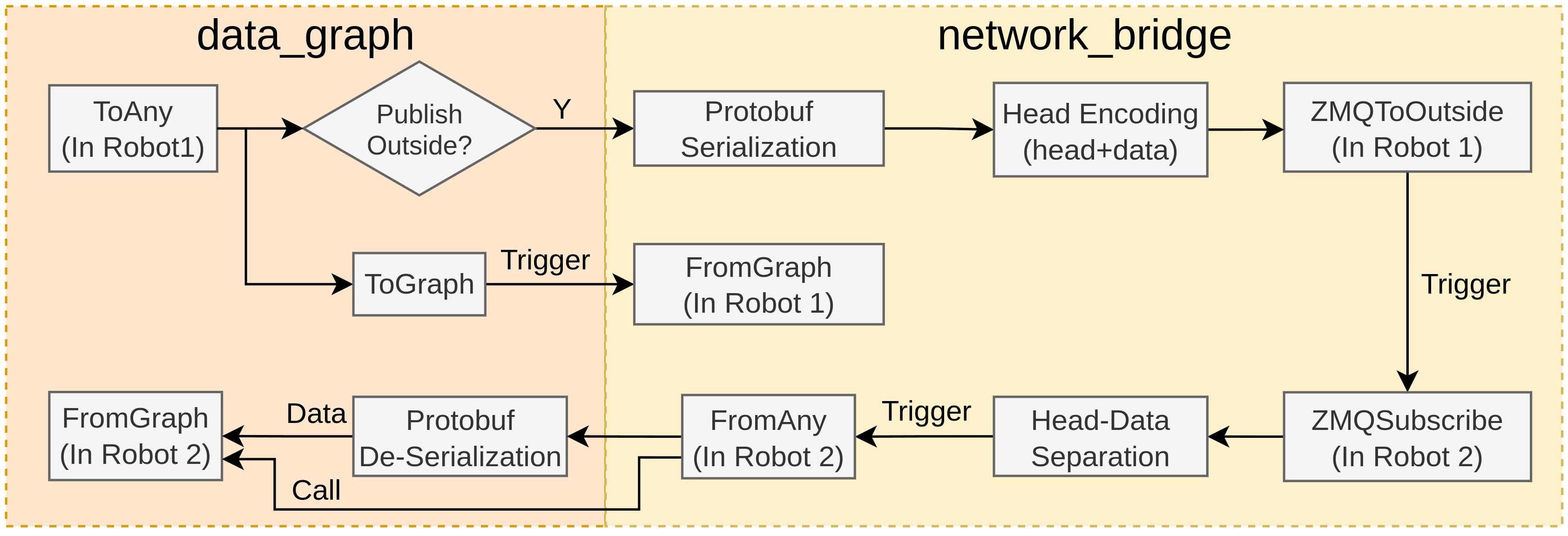}
    \caption{Flowchart of the Communication and Parallel Computing Component. This figure illustrates the detailed process of transmitting data from \textbf{Robot1} to \textbf{Robot2}.}
    \label{communication_pic}
\end{figure*}

\textbf{Lightweight Design}: By separating out less frequently used features, the core library of HyperGraph ROS is only 12 MB. In contrast, the minimal installation of ROS 1 requires at least 200 MB, and ROS 2 requires at least 300 MB. This compact size allows our library to be deployed in resource-constrained environments, thereby significantly conserving storage space.

\textbf{Efficiency first}: HyperGraph ROS prioritizes efficiency over built-in security by using ZeroMQ. This is because DDS alone is insufficient for enterprise security, adds significant overhead, and often requires additional security policies. This approach separates efficiency from security, thereby minimizing redundancy.

\textbf{Distributed Architecture}: Similar to ROS 2, HyperGraph ROS eliminates the need to manually launch a master node. Inter-process communication on the same device occurs directly between nodes, without relying on a central broker.

\textbf{Reducing Learning and Maintenance Costs}: HyperGraph ROS streamlines intra-process, inter-process, and cross-device communication through a unified API that also automates parallel scheduling. This design minimizes user learning curves and maintenance efforts. Additionally, other components are optimized to expose only essential interfaces which further simplifies system interaction.

\textbf{Asynchronous Execution}: HyperGraph ROS leverages TBB's \texttt{function\_node} to facilitate asynchronous execution within robotic systems. The \texttt{function\_node} in TBB operates in two modes: \textit{serial} and \textit{concurrent}. In serial mode, the node processes one message at a time, ensuring that tasks are executed sequentially. In concurrent mode, multiple messages are processed simultaneously, allowing parallel task execution. This flexibility enables HyperGraph ROS to adapt to various computational requirements, thereby enhancing system responsiveness and performance.

\textbf{Cross-Language Support}: By utilizing Protocol Buffers (protobuf) for communication serialization, our system supports multiple programming languages, including Python, C++, C\#, C, Go, Java, and PHP. This approach offers broader language compatibility and higher efficiency compared to ROS's native serialization method.

\textbf{Cross-Platform Support}: HyperGraph ROS is compatible with multiple hardware architectures, including x86, AMD, and ARM. Additionally, it supports various operating systems, such as Windows, Linux, macOS, and Android. This enables deployment on both desktop and embedded platforms, meeting diverse product development needs.

\subsection{Components Overview}
\subsubsection{Communication and Parallel Computing Component}
This component comprises two parts: \textbf{data\_graph} and \textbf{network\_bridge}, as depicted in Fig. \ref{communication_pic}. When \textbf{Robot1} calls \textbf{ToAny}, it first invokes \textbf{ToGraph} to send data within the process of \textbf{Robot1}. Then, based on the YAML configuration file, it determines whether to send the data externally. If the data is to be transmitted externally, it undergoes \textbf{protobuf serialization} to generate a \texttt{data\_string}, which is then combined with the \texttt{head} to form a complete data packet for transmission. The \texttt{head} consists of the following components:

\begin{itemize}
    \item \textbf{header\_length}: A 32-bit field indicating the length of the header, used for extracting the header during unpacking.
    \item \textbf{header\_separator}: A 16-bit separator used to distinguish the header.
    \item \textbf{head\_str}: A string generated from the header's proto, containing channel information.
\end{itemize}

Subsequently, the \texttt{head} and \texttt{data\_string} are transmitted via ZeroMQ to an external network, with the specific destination configured in the YAML file. If the data reaches \textbf{Robot2}, it is first unpacked. If an unpacking error occurs, the packet is discarded. Upon successful unpacking, the \texttt{channel\_name} information and \texttt{data\_string} are extracted, triggering \textbf{FromAny}. 

Since \textbf{FromAny} receives a template, it utilizes the corresponding deserialization function to process the \texttt{data\_string} and then calls \textbf{FromGraph} to deliver the data to the appropriate node in \textbf{Robot2}.

\subsubsection{Transformation Tree Component}
\begin{figure}[thpb]
    \centering
    \vspace{0.1cm} 
    \includegraphics[scale=0.1]{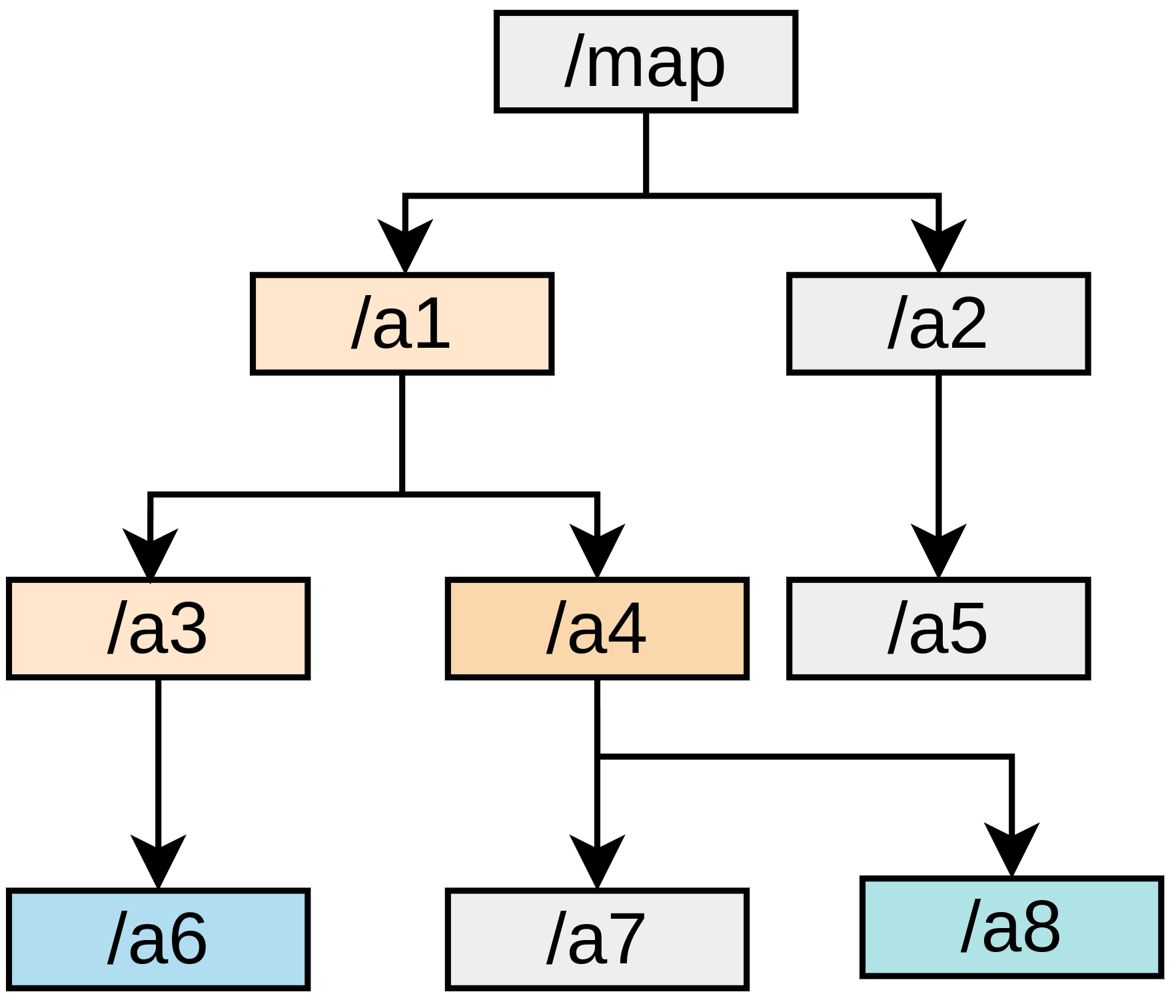}
    \caption{Illustration of the transformation\_tree. When querying the transformation matrix between the blue nodes \texttt{/a6} and \texttt{/a8}, the algorithm first identifies their common ancestor \texttt{/a1}. It then computes the transformation matrices from \texttt{/a6} to \texttt{/a1} and from \texttt{/a1} to \texttt{/a8}, whose product yields the transformation matrix between \texttt{/a6} and \texttt{/a8}.}
    \label{tf_tree}
\end{figure}

This component stores the coordinate transformations between different sensors and maps. Notably, instead of relying on existing libraries, we implement a transformation tree manually. When querying the transformation between two coordinate frames, the system identifies their common ancestor and computes the relative transformation along the chain, as illustrated in Fig. \ref{tf_tree}. This implementation is more lightweight and easier to maintain.

\subsubsection{Parameter Server Component}
This component implements a parameter server utilizing YAML~\cite{benkiki_yaml_2009} files for configuration management. The parameter server serves as a centralized storage for configuration parameters, enabling nodes to retrieve and set parameters at runtime. By leveraging YAML—a human-readable data serialization standard—the system facilitates easy configuration and maintenance. Parameters are defined in YAML files and loaded into the server at startup, allowing dynamic updates without recompiling the code. This design enhances flexibility and simplifies the management of complex configurations within the robotic system.

\begin{figure*}[thpb]
    \centering
    \includegraphics[scale=0.15]{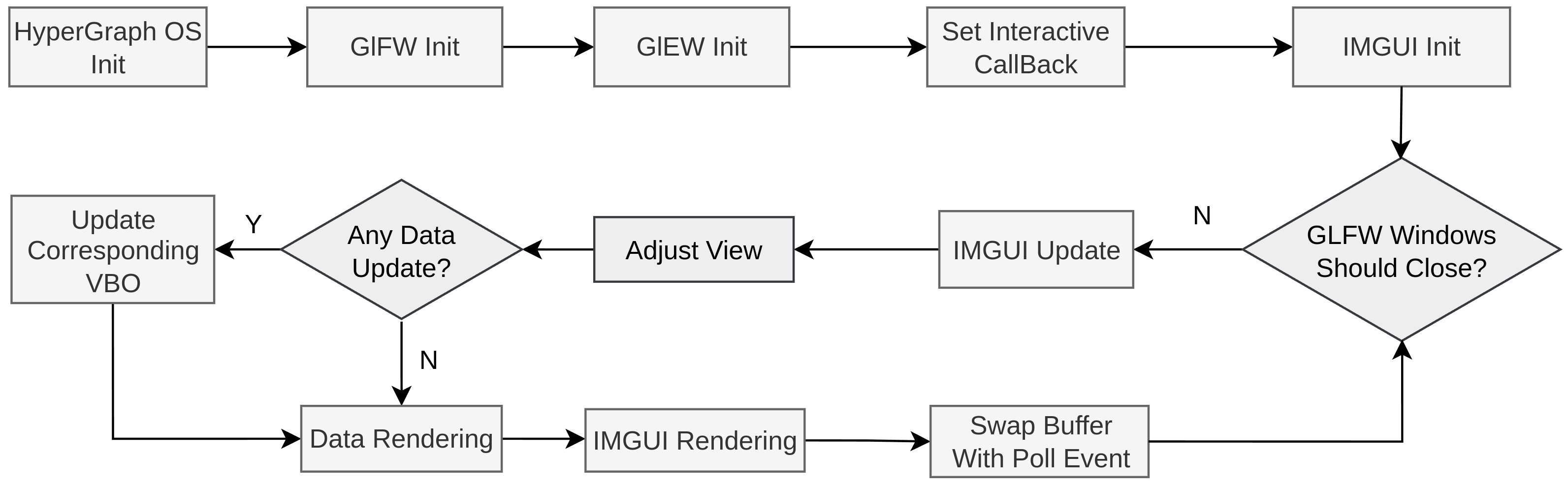}
    \caption{Flowchart of the Visualization Component. It illustrates the rendering and triggering logic of our UI.}
    \label{ui_construct}
\end{figure*}

\subsubsection{Time System Component}
This component is designed to facilitate time synchronization during the execution of various robotic algorithms in both simulation and real-world scenarios. It allows the system to set an initial reference time at startup, after which any calls to its interface return the relative time elapsed since this reference point. This functionality ensures consistent temporal coordination across different modules, enhancing the reliability and predictability of robotic operations.

\subsubsection{Visualization Component}
This component is designed for efficient visualization of intermediate algorithmic results, such as point clouds and pose trajectories. Developed as a standalone project, independent of the aforementioned components, it receives messages on specified topics via the Communication and Parallel Computing Component's API. The detailed workflow is illustrated in Fig. \ref{ui_construct}.

The initialization process begins with HyperGraph ROS, followed by GLFW and FLEW. Interactive callbacks are then configured for mouse buttons, the scroll wheel, mouse movement, and keyboard inputs, enabling free-view navigation and zoom operations. IMGUI is subsequently initialized before entering the GLFW loop.

Each iteration of the loop starts with an IMGUI update. The system then adjusts the view based on the latest interaction parameters and determines whether a data update is required. If the \textbf{FromAny} function updates the data—indicating that visualization data has been received from another process—a corresponding update flag is set, triggering a VBO update. To enhance rendering efficiency, VBO updates occur only when necessary.

Finally, IMGUI and data rendering are performed, followed by buffer swapping and event polling before the next iteration. This component avoids reliance on heavy libraries like Qt, achieving significantly faster rendering speeds and lower memory consumption. Furthermore, its foundation on Protocol Buffers (protobuf) extends compatibility beyond ROS, thereby enabling broader cross-language visualization support.

\section{Use Cases}
This section presents various use cases encountered during robot system development and proposes solutions based on our operating system.

\subsection{Use Case 1: Publish and Subscribe message}
\subsubsection{Common Usage}

We employ a publisher-subscriber model for message exchange. To transmit information, users can utilize the \textbf{ToAny} interface in a non-blocking manner, with the format:

\begin{lstlisting}[breaklines=true, numbers=none, xleftmargin=7pt, xrightmargin=7pt]
    ToAny(channel_name, data);
\end{lstlisting}

Messages are buffered within the TBB flow graph. This function dispatches data to the \textbf{FromAny} callback functions associated with the specified \textbf{channel\_name}.

For receiving messages, the \textbf{FromAny} interface is used as follows:

\begin{lstlisting}[breaklines=true, numbers=none, xleftmargin=7pt, xrightmargin=7pt]
    FromAny<template_name>(channel_name, invoke_type, callback);
\end{lstlisting}

Here, \textbf{template\_name} denotes the type of the received message, essential for deserialization. The \textbf{invoke\_type} parameter specifies whether messages are queued for sequential execution or processed concurrently. The \textbf{callback} function is invoked upon receiving a message, and is automatically scheduled and load-balanced by Intel-TBB. 

User-defined data types must implement serialization and deserialization methods or be generated by Protocol Buffers (protobuf) to employ the aforementioned APIs. In the absence of such methods, simplified APIs are available:

\begin{lstlisting}[breaklines=true, numbers=none, xleftmargin=7pt, xrightmargin=7pt]
    ToGraph(channel_name, data);
\end{lstlisting}

\begin{lstlisting}[breaklines=true, numbers=none, xleftmargin=7pt, xrightmargin=7pt]
    FromGraph<template_name>(channel_name, invoke_type, callback);
\end{lstlisting}

However, these simplified interfaces are limited to intra-process communication and cannot transmit data across processes.

\subsubsection{Inter-Process Communication and Transmission Control}

\begin{figure}[thpb]
    \centering
    \includegraphics[width=0.45\textwidth]{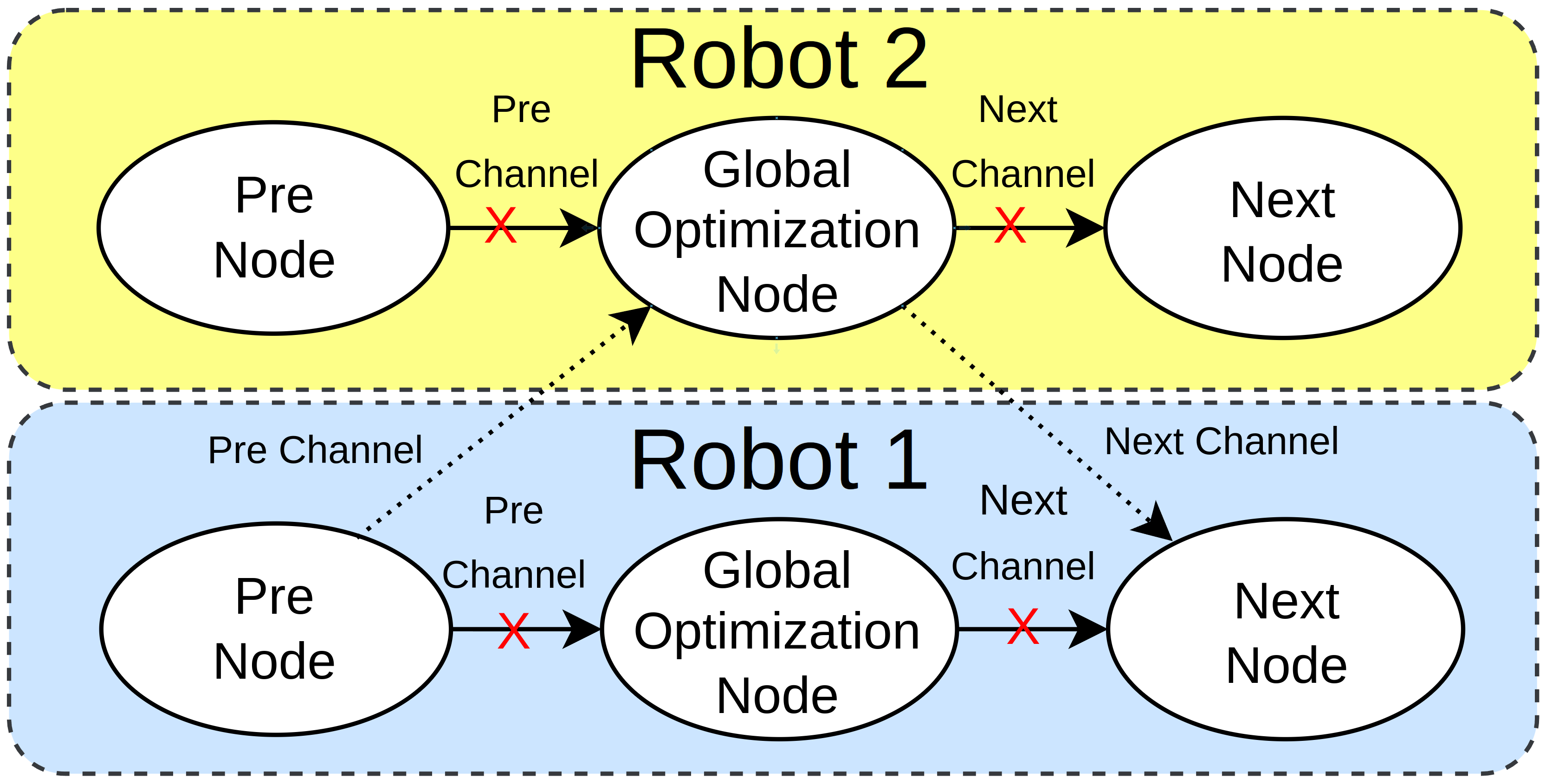}
    \caption{Cross-Device Communication Case. The Global Optimization Node of Robot1 is deployed on Robot2 during operation, and the results are transmitted back to Robot1.}
    \label{network_case}
\end{figure}

In practical robotic systems, such as UAV-UGV collaboration and cloud-based mapping, certain modules, such as global optimization, may need to run on cloud servers or other robots, as shown in Fig. \ref{network_case}.

This can be configured in \textbf{Robot 1} via a YAML file as shown in Listing \ref{listing_network} and Listing \ref{listing_node}. A similar configuration can be applied to Robot2.
\begin{figure}[H] 
\begin{lstlisting}[caption={Configuration of \texttt{network\_setting.yaml}}, 
label={listing_network}, breaklines=true, basicstyle=\ttfamily\small, numbers=none, 
xleftmargin=7pt, xrightmargin=7pt]
network:
    publisher:
        ip: "tcp://*:5553"
        channels:
            - pre_channel
    subscribers:
        - ip: "x"
        channels:
            - next_channel
\end{lstlisting}
\end{figure}

\begin{lstlisting}[caption={Configuration of \texttt{data\_graph.yaml}}, label={listing_node}, breaklines=true, basicstyle=\ttfamily\small, numbers=none, xleftmargin=7pt, xrightmargin=7pt]
    network:
      publisher:
        ip: "tcp://*:5553"
        channels:
          - pre_channel
          - next_channel
\end{lstlisting}

This configuration allows for flexible deployment of computational modules across different hardware resources, optimizing performance and resource utilization.

\subsection{Use Case 2: Visualization}
A straightforward approach is to launch the visualization program and use \textbf{ToAny} to publish the corresponding topics, which can be configured in the YAML file. For instance, topics such as \textbf{lidar} and \textbf{re\_optimized\_poses} can be transmitted, while \texttt{network\_setting.yaml} can be set as Listing \ref{listing_config_visualization}. 

\begin{lstlisting}[caption={Configuration of \texttt{network\_setting.yaml}}, label={listing_config_visualization}, breaklines=true, basicstyle=\ttfamily\small, numbers=none, xleftmargin=7pt, xrightmargin=7pt]
    network:
      publisher:
        ip: "tcp://*:5553"
        channels:
          - lidar_visual_channel
          - re_optimized_poses
\end{lstlisting}

\begin{figure}[thpb]
    \centering
    \includegraphics[scale=0.106]{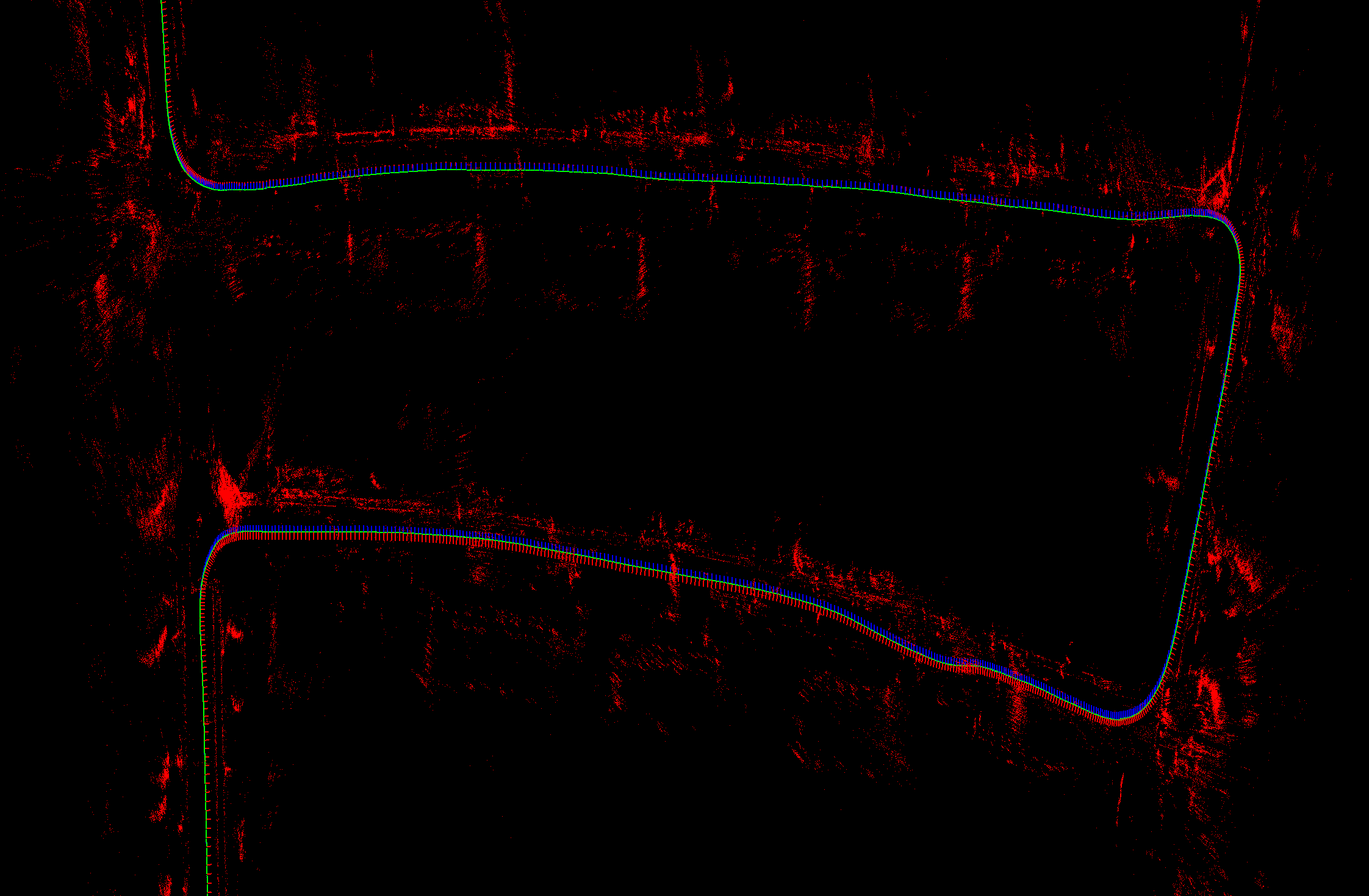}
    \caption{One Example of Visualization. The figure illustrates the simulation results of a simultaneously localization and mapping (SLAM) process. The red points represent the point cloud, while the trajectory is visualized using an \texttt{axes} representation (similar to ROS RViz), which consists of a coordinate frame with three axes: x, y, and z. Multiple overlapping frames form the axes trajectory.}
    \label{ui_exp}
\end{figure}

\begin{figure}[t]
    \centering
    \includegraphics[scale=0.05]{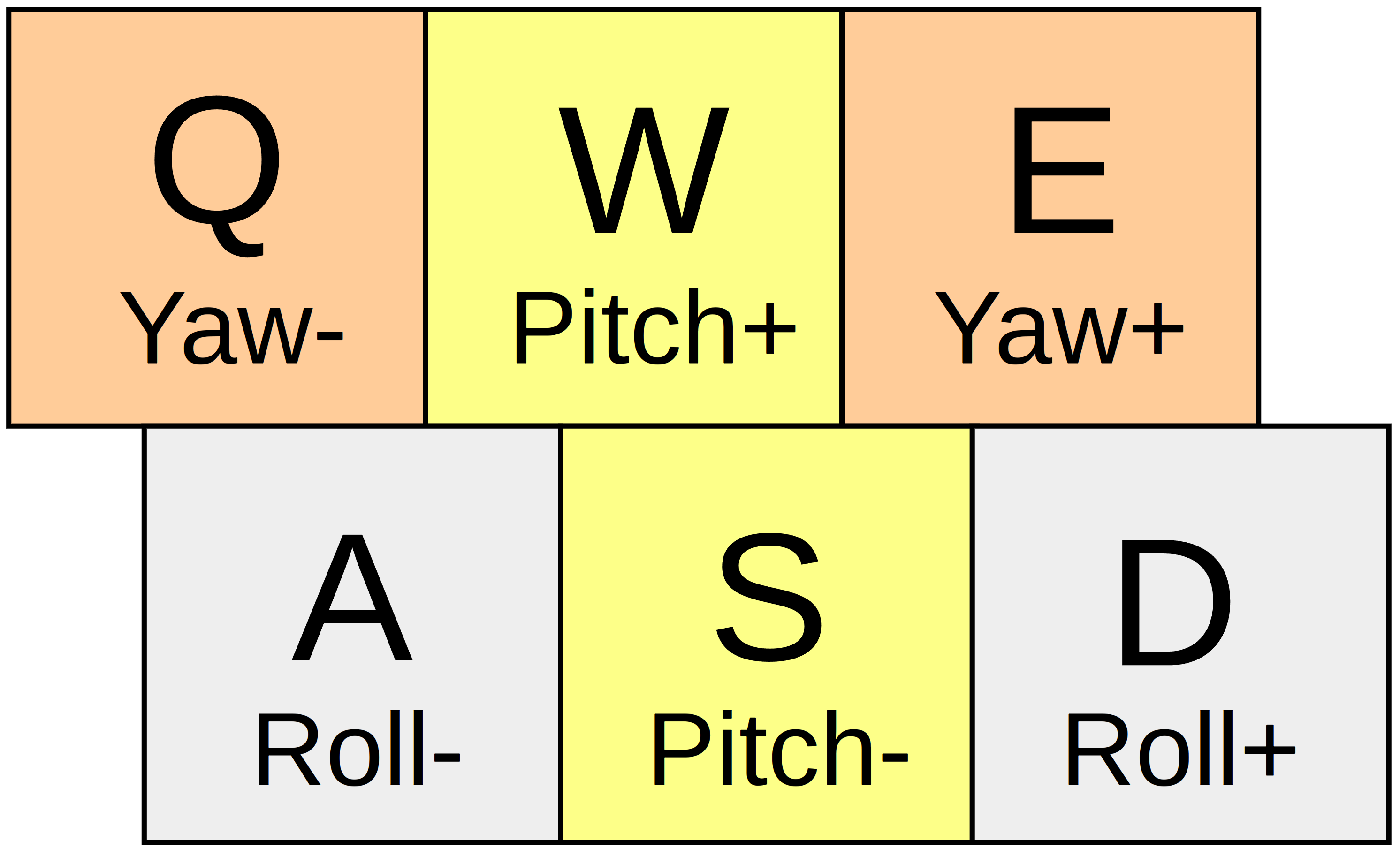}
    \caption{Illustration of Euler angle adjustment for viewpoint control using the keyboard.}
    \label{key_pic}
\end{figure}

It also supports free viewpoint adjustment, which includes the following functionalities:

\begin{enumerate}
    \item \textbf{Zooming}: Press the left mouse button in the UI to select a zoom anchor point, then scroll the mouse wheel to zoom in or out.
    \item \textbf{Panning}: Hold the right mouse button and drag the mouse to pan the view.
    \item \textbf{Viewpoint Adjustment}: Click the left mouse button in the UI to select a viewpoint anchor, then use the keyboard for fine-tuned adjustments, as illustrated in Fig. \ref{key_pic}.
\end{enumerate}

This visualization software supports real-time rendering, with a visualization example illustrated in Fig. \ref{ui_exp}.

\section{Experimental Evaluation}
\label{sec:experiments}

This section evaluates the communication performance and visualization efficiency of HyperGraph ROS compared to ROS and ROS 3D Robot Visualizer (RViz). The communication experiments evaluate intra-process, inter-process, and cross-device communication in terms of latency, while throughput is assessed only for intra-process communication. The visualization experiment assesses CPU and memory usage in HyperGraph ROS VIZ versus ROS~2 RViz. All tests were conducted on a laptop with an Intel i7-10750H processor, 32GB RAM, and an NVIDIA RTX 2060 GPU running Ubuntu 20.04 LTS. For cross-device communication, an additional desktop with an Intel i5-10400F processor, 16GB RAM, and an NVIDIA GTX 1050 GPU was used as the message receiver.

\subsection{Communication Latency Performance}
\label{subsec:communication_performance}

\subsubsection{Intra-process Communication}
\label{subsubsec:intra_process_communication}
In intra-process communication, \textbf{HyperGraph ROS} utilizes \texttt{Intel-TBB} and direct pointer passing for efficient data exchange, while \textbf{ROS 1} employs the \texttt{Nodelet} mechanism to enable shared memory communication, reducing message copying. \textbf{ROS 2} uses \texttt{Intra-Process Communication (IPC)} by default (non-optimized ROS 2), achieving zero-copy transfer with the \texttt{use\_intra\_process\_comms} flag enabled (optimized ROS 2).

\begin{figure}[thpb]
    \centering
    \includegraphics[scale=0.106]{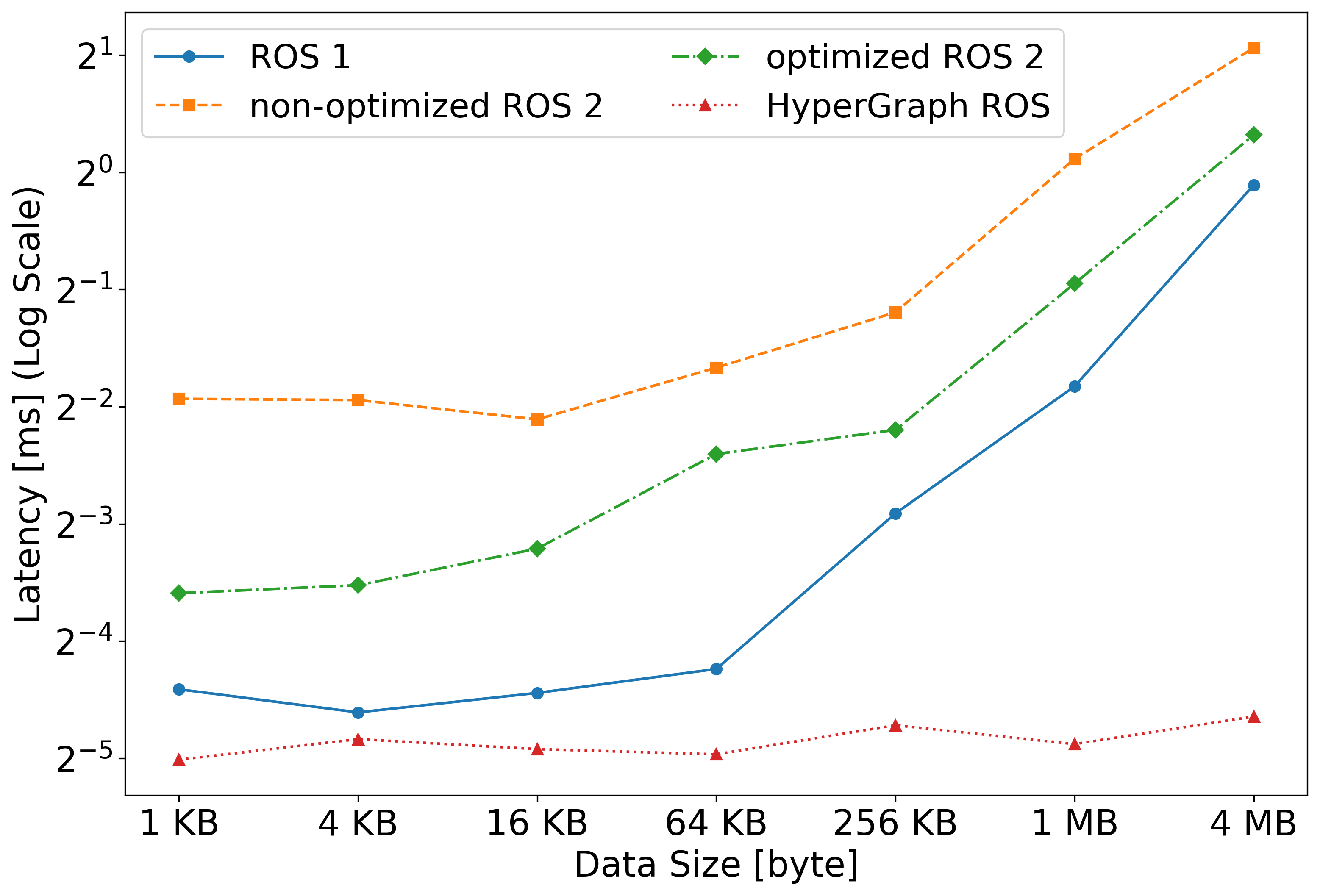}
    \caption{Average latency per message in intra-process communication for ROS 1, optimized ROS 2, non-optimized ROS 2, and HyperGraph ROS.}
    \label{intra-process_latency}
\end{figure}

This experiment evaluates latency by transmitting \textbf{100} data packets of varying sizes (1KB to 4096KB). The average latency is shown in Fig.~\ref{intra-process_latency}.
Since HyperGraph ROS utilizes pure pointer passing, its performance is constrained only by hardware limitations, making it superior to other operating systems regardless of data size.

\begin{figure}[thpb]
    \centering
    \includegraphics[scale=0.106]{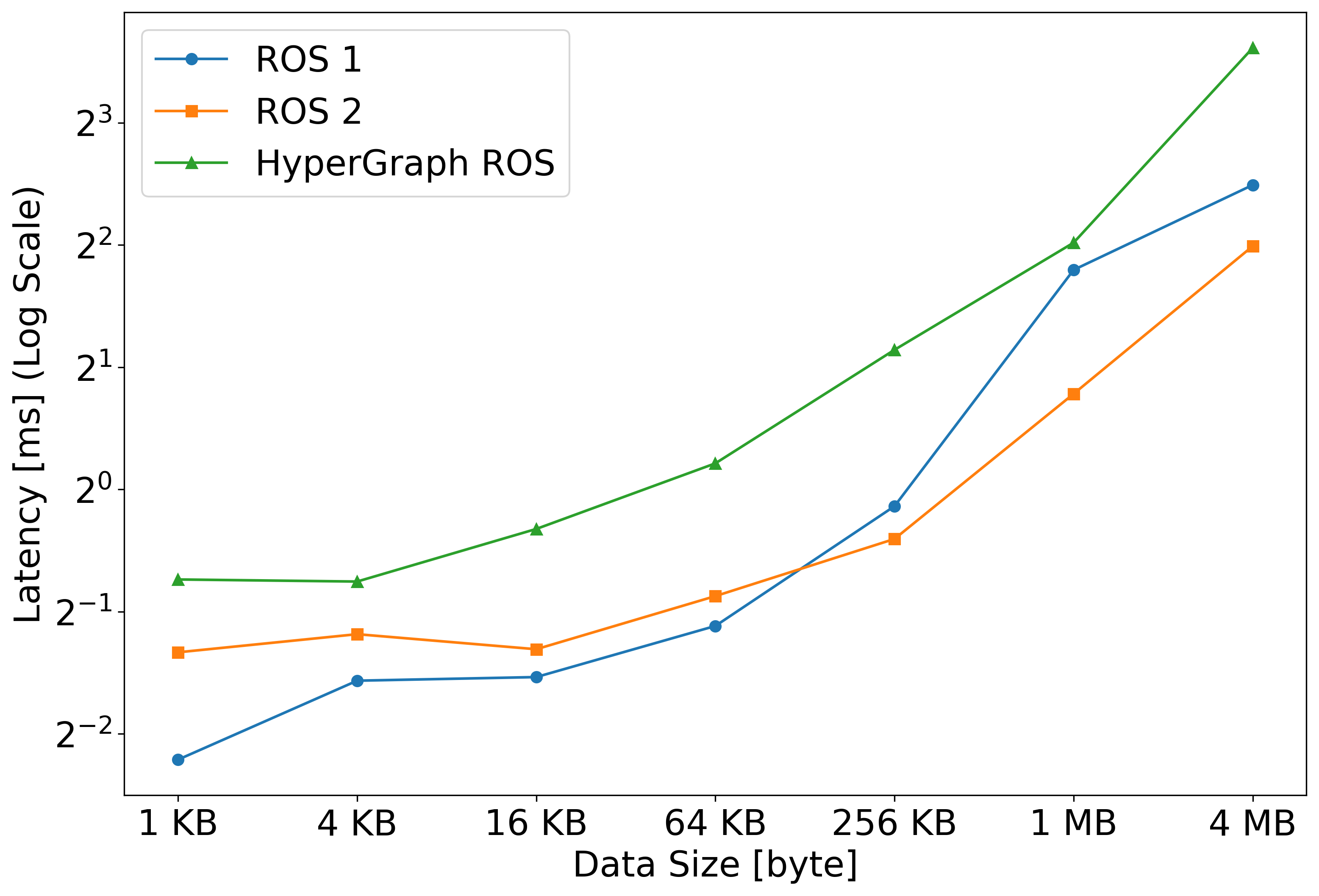}
    \caption{Average latency per message in inter-process communication for ROS 1, ROS 2, and HyperGraph ROS.}
    \label{Inter-process_latency}
\end{figure}

\subsubsection{Inter-process Communication}
\label{subsubsec:inter_process_communication}
For inter-process communication, \textbf{HyperGraph ROS} uses \texttt{ZeroMQ (IPC Mode)} for message exchange, while \textbf{ROS 1} relies on \texttt{TCPROS/UDPROS} with a centralized \texttt{ROS Master}, and \textbf{ROS 2} employs \texttt{Data Distribution Service (DDS)} middleware. In the experiment, Publisher and Subscriber nodes ran on the same computer to measure latency differences under different data loads (same as \ref{subsubsec:intra_process_communication}). The results are shown in Fig.~\ref{Inter-process_latency}.
Our system exhibits slightly lower efficiency than ROS~1 and ROS~2 in inter-process communication, as we employ serialization isolation before transmitting via shared memory. This approach enhances data integrity, prevents unintended memory modifications, and improves compatibility across heterogeneous processes, albeit with added serialization overhead. Future work will focus on reducing latency while maintaining data safety.



\subsubsection{Cross-device Communication}
\label{subsubsec:cross_device_communication}

\begin{figure}[t]
    \centering
    \includegraphics[scale=0.106]{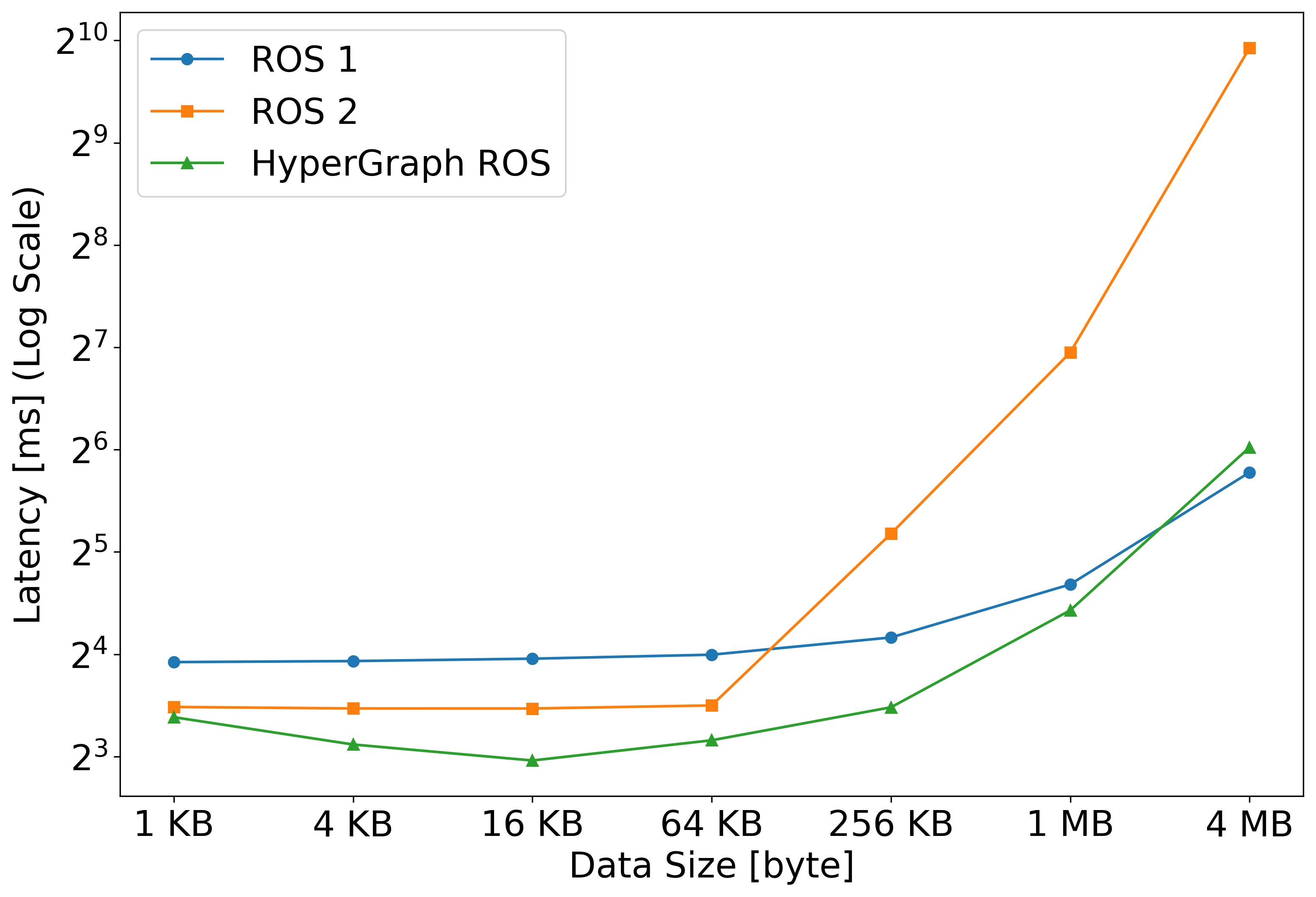}
    \caption{Average latency per message in cross-device communication ROS 1, ROS 2, and HyperGraph ROS in Cross-device Communication.}
    \label{Cross-device_latency}
\end{figure}

In cross-device communication experiments, \textbf{HyperGraph ROS} utilizes \texttt{ZeroMQ}, \textbf{ROS 1} employs \texttt{TCPROS/UDPROS}, and \textbf{ROS 2} leverages \texttt{DDS}. Publisher and Subscriber nodes were deployed on separate devices, following the same parameters and procedures as in Section~\ref{subsubsec:intra_process_communication}. The final results of communication latency are presented in Fig.~\ref{Cross-device_latency}.
It can be seen that our system is comparable to ROS 1 in terms of communication latency. However, compared to ROS 2, our system has a significant advantage. As the data size increases, the latency per message of ROS 2 rises considerably, while that of our system increases only slightly. For example, when the data size is 4MB, the transmission latency of ROS 2 is approximately 970ms, whereas our system has a latency of less than 80ms. This substantial difference demonstrates the high efficiency of our system in cross-device communication.

\subsection{Communication Throughput Performance}
\label{subsec:communication_throughput_performance}
\begin{figure}[t]
    \centering
    \includegraphics[scale=0.106]{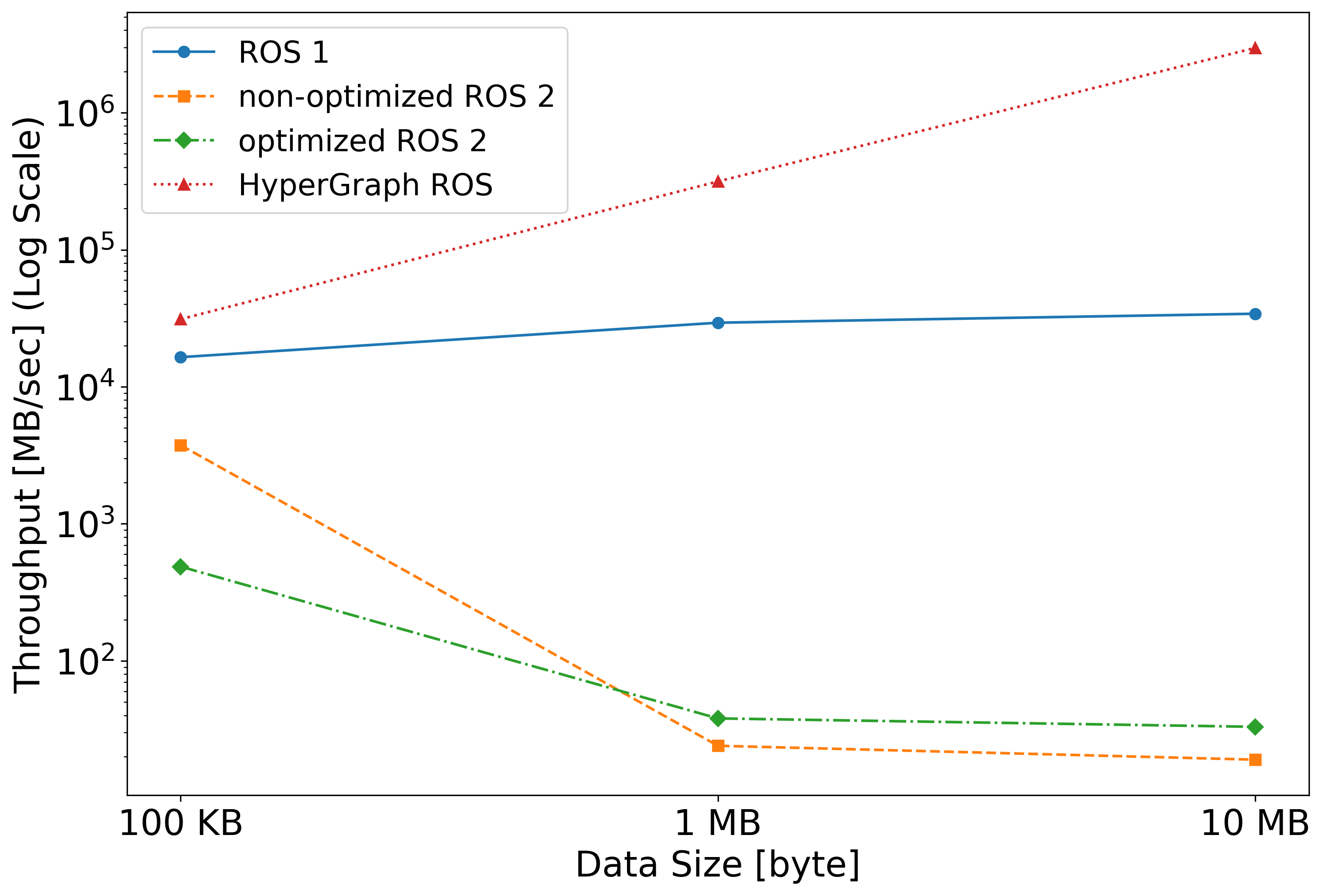}
    \caption{Throughput comparison in intra-process communication for ROS 1, non-optimized ROS 2, optimized ROS 2, and HyperGraph ROS.}
    \label{intra-process_throughput}
\end{figure}

The Communication Throughput Performance experiment primarily focuses on intra-process communication throughput. This is because the HyperGraph in HyperGraph ROS is constructed from an intra-process Intel-TBB Graph interconnected by inter-process ZMQ edges. Consequently, the overall performance of HyperGraph is largely determined by intra-process efficiency. Therefore, an additional intra-process throughput experiment is conducted in this section. Specifically, we evaluate throughput by continuously transmitting data ranging from 100KB to 10MB at a frequency of 10Hz. The final results are presented in Fig.~\ref{intra-process_throughput}.

Throughput is calculated by summing the transmission latency of each data packet to obtain the total transmission time and dividing the total transmitted data size by this time. This method ensures a more precise estimation of the theoretical upper bound of throughput performance.

As shown in Fig.~\ref{intra-process_throughput}, our system demonstrates significantly higher intra-process communication throughput due to Intel TBB's parallelized reception mechanism.

\subsection{Visualization Efficiency}
\label{subsec:visualization_efficiency}

\begin{figure}[t]
    \centering
    \includegraphics[scale=0.106]{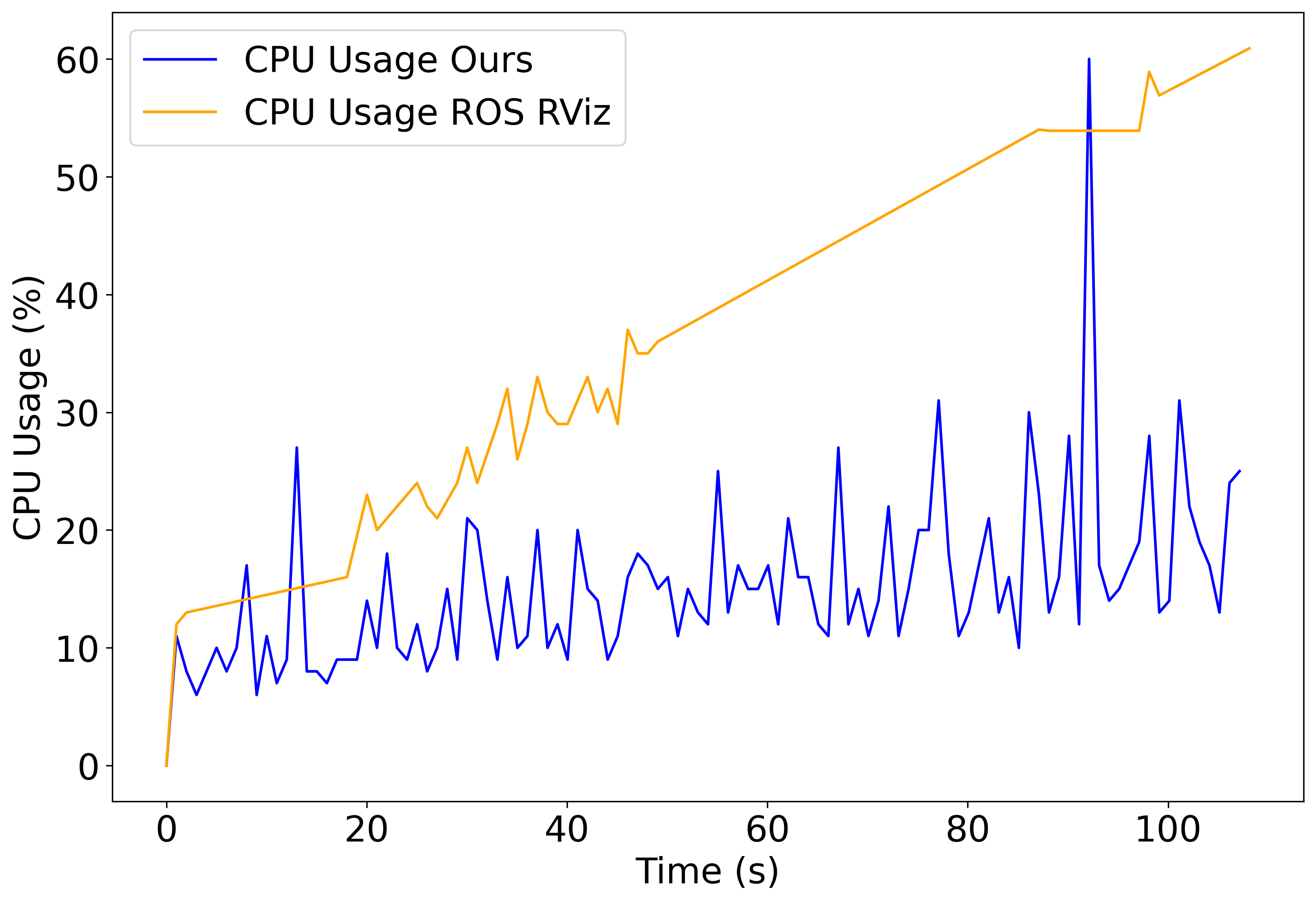}
    \caption{Comparison of CPU usage between HyperGraph ROS VIZ and ROS RViz.}
    \label{T_cloud cpu usage}
\end{figure}

\begin{figure}[t]
    \centering
    \includegraphics[scale=0.106]{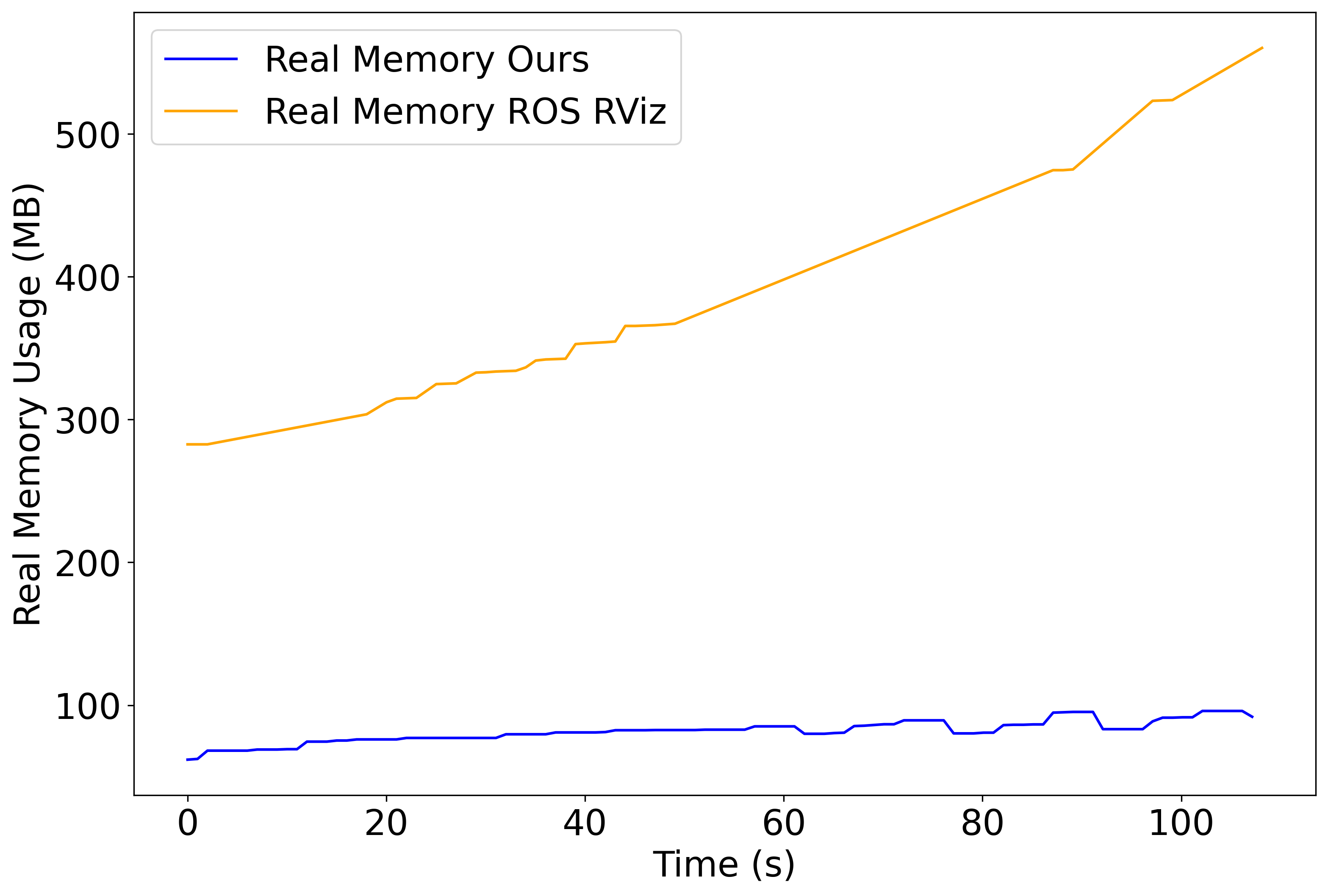}
    \caption{Comparison of memory between HyperGraph ROS VIZ and ROS RViz.}
    \label{T_cloud cpu real mamory}
\end{figure}

In terms of visualization, \textbf{HyperGraph ROS VIZ}, built on \textbf{OpenGL} and \textbf{IMGUI}, outperforms \textbf{ROS RViz} (which uses \textbf{OGRE3D} and \textbf{Qt}) in CPU and memory efficiency, as tested on an Intel i7-12700H laptop with 16GB RAM using SLAM trajectory and point cloud data (Fig. \ref{ui_exp}), with CPU usage monitored in Fig. \ref{T_cloud cpu usage}. From this figure, it can be seen that as time progresses, the CPU usage of ROS RViz increases linearly, while the CPU usage of our visualization tool does not show significant growth. Moreover, from Fig.~\ref{T_cloud cpu real mamory}, we can see that our visualization tool is also very stable in terms of memory consumption, with almost no increase over time. In contrast, the memory usage of ROS RViz increases significantly.

\section{Conclusions}
\label{conclusion_sec}
In this paper, we introduced \textbf{HyperGraph ROS}, an open-source robotic operating system designed for efficient hybrid parallel computing through a computational hypergraph. By integrating Intel-TBB for dynamic parallelism and ZeroMQ for high-performance messaging, our system achieves lower transmission latency and higher throughput compared to ROS 1 and ROS 2. Additionally, the lightweight visualization tool significantly improves rendering efficiency while minimizing CPU and memory usage. These features make HyperGraph ROS a promising solution for real-time and resource-constrained robotic applications.

\bibliographystyle{IEEEtranN}  
\bibliography{IEEBI}  

\end{document}